\pdfoutput=1

\documentclass[11pt]{article}

\usepackage[]{acl}

\usepackage{times}
\usepackage{latexsym}
\usepackage{amsmath,amssymb,amsfonts}
\usepackage{graphicx}
\usepackage{tabularx}
\usepackage{multirow}
\usepackage{longtable}
\usepackage{booktabs}
\usepackage{subfigure}
\usepackage{algpseudocode}
\usepackage[ruled,vlined]{algorithm2e}
\include{pythonlisting}
\usepackage{subfiles}
\usepackage{makecell}
\usepackage{dblfloatfix}    
\makeatletter
\DeclareRobustCommand{\cev}[1]{%
  \mathpalette\do@cev{#1}%
}
\newcommand{\do@cev}[2]{%
  \fix@cev{#1}{+}%
  \reflectbox{$\m@th#1\vec{\reflectbox{$\fix@cev{#1}{-}\m@th#1#2\fix@cev{#1}{+}$}}$}%
  \fix@cev{#1}{-}%
}
\newcommand{\fix@cev}[2]{%
  \ifx#1\displaystyle
    \mkern#23mu
  \else
    \ifx#1\textstyle
      \mkern#23mu
    \else
      \ifx#1\scriptstyle
        \mkern#22mu
      \else
        \mkern#22mu
      \fi
    \fi
  \fi
}

\usepackage{pifont}
\newcommand{\cmark}{\ding{51}}%
\newcommand{\xmark}{\ding{55}}%
\usepackage{fontawesome}

\usepackage{array}
\newcolumntype{L}[1]{>{\raggedright\let\newline\\\arraybackslash\hspace{0pt}}m{#1}}
\newcolumntype{C}[1]{>{\centering\let\newline\\\arraybackslash\hspace{0pt}}m{#1}}
\newcolumntype{R}[1]{>{\raggedleft\let\newline\\\arraybackslash\hspace{0pt}}m{#1}}

\usepackage[T1]{fontenc}

\usepackage[utf8]{inputenc}

\usepackage{microtype}
\usepackage{inconsolata}
\title{Deciphering Hate: Identifying Hateful Memes and Their Targets}

\author{Eftekhar Hossain{\textsuperscript{$\clubsuit$}}, Omar Sharif{\textsuperscript{$\spadesuit$}}, Mohammed Moshiul Hoque{\textsuperscript{\faCny}}, Sarah M. Preum{\textsuperscript{$\spadesuit$}}\\
   {\textsuperscript{\faCny}}Department of Computer Science and Engineering\\ 
  {\textsuperscript{$\spadesuit$}}Department of Computer Science, Dartmouth College, USA \\
{\textsuperscript{$\clubsuit$}}Department of Electronics and Telecommunication Engineering\\ 
  {\textsuperscript{$\clubsuit$ \faCny}}Chittagong University of Engineering \& Technology, Bangladesh \\
 \texttt{\small \{eftekhar.hossain, moshiul\_240\}@cuet.ac.bd}, \texttt{\small\{omar.sharif.gr, sarah.masud.preum\}@dartmouth.edu}} 

\begin{document}
\maketitle
\begin{abstract}
Internet memes have become a powerful means for individuals to express emotions, thoughts, and perspectives on social media. While often considered a source of humor and entertainment, memes can also disseminate hateful content targeting individuals or communities. 
Most existing research focuses on the negative aspects of memes in high-resource languages, overlooking the distinctive challenges associated with low-resource languages like Bengali (also known as Bangla). Furthermore, while previous work on Bengali memes has focused on detecting hateful memes, there has been no work on detecting their targeted entities. To bridge this gap and facilitate research in this arena, we introduce a novel multimodal dataset for Bengali, \textbf{BHM} (Bengali Hateful Memes). The dataset consists of 7,148 memes with Bengali as well as code-mixed captions, tailored for two tasks: \textit{(i) detecting hateful memes}, and \textit{(ii) detecting the social entities they target (i.e., Individual, Organization, Community, and Society)}. To solve these tasks, we propose {\fontfamily{qcr}\selectfont DORA} (\underline{D}ual c\underline{O}-attention f\underline{RA}mework), a multimodal deep neural network that systematically extracts the significant modality features from the memes and jointly evaluates them with the modality-specific features to understand the context better.  Our experiments show that {\fontfamily{qcr}\selectfont DORA} is generalizable on other low-resource hateful meme datasets and outperforms several state-of-the-art rivaling baselines.  

\end{abstract}
\section{Introduction}
In recent years, social media has brought a distinct form of multimodal entity: \textit{memes}, providing a means to express ideas and emotions. Memes are compositions of images coupled with concise text. While memes are often amusing, they can also spread hate by incorporating socio-political elements. These hateful memes pose a significant threat to social harmony as they have the potential to harm individuals or specific groups based on factors like political beliefs, sexual orientation, or religious affiliations. Despite the significant influence of memes, their multifaceted nature and concealed semantics make them very hard to analyze. The prevalence of highly toxic memes in recent times has led to a growing body of research into the negative aspects of memes, such as hate \cite{kiela2020hateful}, offensiveness \cite{shang2021aomd}, and harm \cite{pramanick2021momenta}. However, most works focused on the memes in high-resource languages, while only a few studied the objectionable (i.e., hate, abuse) memes of low-resource languages \cite{kumari2023emoffmeme}. This is particularly true for Bengali.
\begin{figure}[t!]
  \centering
  \subfigure{\includegraphics[scale=0.6,height = 3.5cm,width =2.8cm]{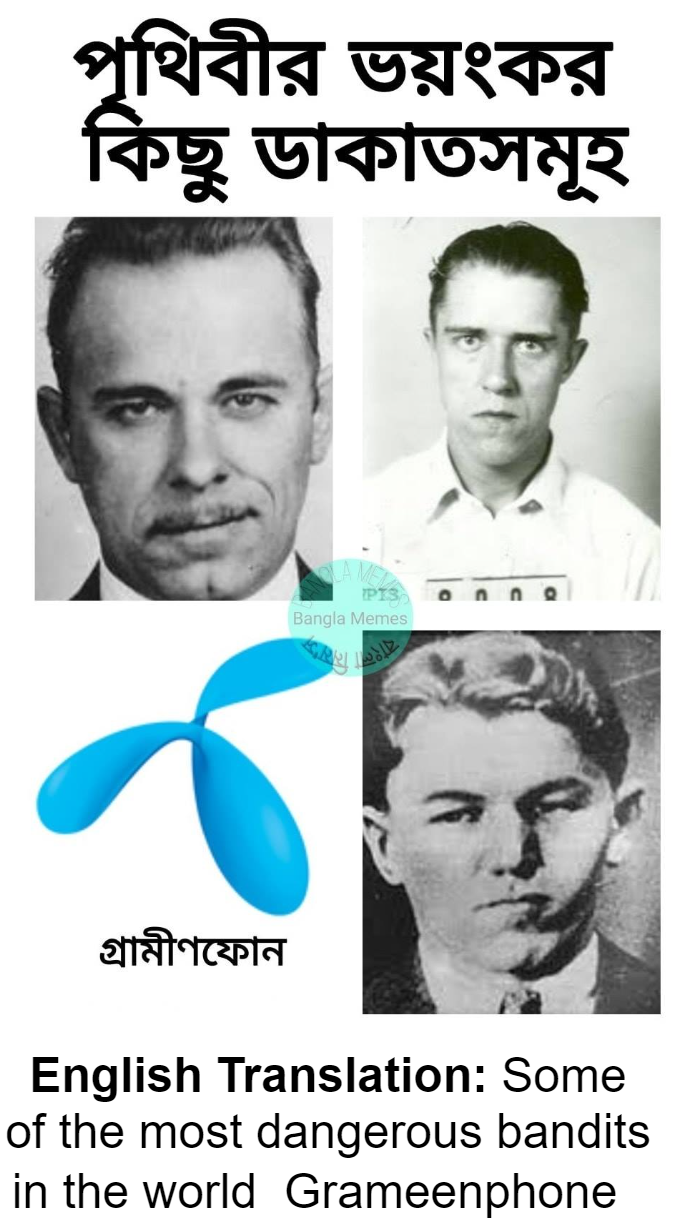}
  }\quad
  \subfigure{\includegraphics[scale=0.6]{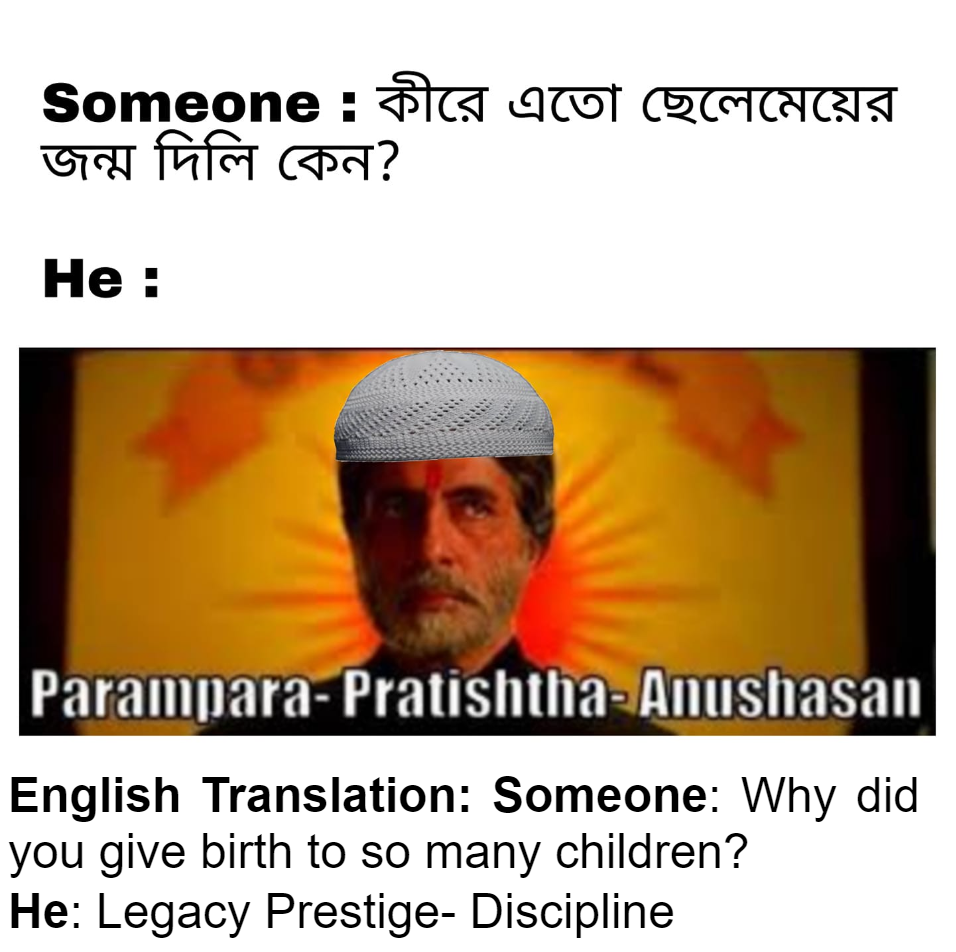}}

 \caption{Example of hateful memes with associated targets. The first meme directly refers to a telecom organization as a bandit, and the second one deliberately attacks a religious community.}
 \label{fig:sample memes}
\end{figure}

Despite being the seventh most widely spoken language, having 210 million speakers globally, Bengali is considered one of the notable resource-constrained languages \cite{das2023banglaabusememe}. Moreover, it is the official language of Bangladesh and holds recognition as one of the official languages in the constitution of India. So, developing resources in Bengali is important to build more inclusive language technologies.  Statistics indicate that over 45 million users engage with Bengali on various social media platforms daily \cite{sharif2022tackling}. Recently, memes have gained significant traction in social media, reaching a broad audience and influencing public sentiment. Many of these memes contain hateful content targeting various social entities. The limited availability of the benchmark datasets primarily constrains the identification of such hateful memes. Only two prior works \cite{karim2022multimodal, hossain-etal-2022-mute} developed hateful meme datasets in Bengali. However, both of them overlook the targets associated with the hateful memes. For instance, the first meme in figure \ref{fig:sample memes} is hateful towards an organization because it depicts a company (i.e., Grameenphone) as a robber. Similarly, the second meme propagates hate towards a specific religious community (i.e., Muslim) by highlighting that they produce many children. Therefore, identifying targets of hateful memes is crucial for 1) understanding targeted groups and developing interventions to counter hate speech and 2) personalizing content filters, ensuring that users are not exposed to hateful content directed at them or their communities.

To bridge this research gap, we develop a novel Bengali meme dataset encompassing the targeted entities of hate. The captions in the dataset contain code-mixed (Bengali + English) and code-switched text (written Bengali
dialects in English alphabets). This makes the dataset more distinctive and challenging compared to previous studies in the field. On the technical front, prior research on hateful meme detection \cite{kiela2019supervised, pramanick2021detecting} revealed that off-the-shelf multimodal systems, which often perform well on a range of visual-linguistic tasks, struggle when applied to memes. Besides, most of the existing visio-linguistic models \cite{radford2021learning,li2019visualbert,li2022blip} are primarily trained on image-text pairs of English languages, thus limiting their capability on low-resource languages. Moreover, the existing state-of-the-art multimodal models \cite{pramanick-etal-2021-momenta-multimodal, lee2021disentangling} for hateful meme detection can not be replicated because several components of their architectures are not available in low-resource languages (i.e., Bengali). To tackle these issues, we developed a multimodal framework, and our major contributions are as follows:
\begin{itemize}
    \item We develop a benchmark multimodal dataset comprising 7,148 Bengali memes. The dataset includes two sets of labels for (i) detecting hateful memes and (ii) identifying targeted entities (individuals, organizations, communities, and society). We also provide detailed annotation guidelines to facilitate resource creation for other low-resource languages in this domain.
    
    \item We propose {\fontfamily{qcr}\selectfont DORA}, a multimodal framework to identify hateful memes and their targets. We also perform extensive experiments on \textbf{BHM} dataset and show that {\fontfamily{qcr}\selectfont DORA} outperforms nine state-of-the-art unimodal and multimodal baselines in terms of all the evaluation measures. We further establish the generalizability and transferability of {\fontfamily{qcr}\selectfont DORA} on two existing benchmark hateful meme datasets in Bengali and Hindi.  
    
\end{itemize}
\section{Related Work}
\textbf{Hateful memes dataset:} Over the past few years, several meme datasets have been developed regarding hate speech and its various intensity levels, such as offense, harm, abuse, and troll. \newcite{sabat2019hate} developed a hateful memes dataset containing 5,020 memes. Facebook AI \cite{kiela2020hateful} contributed to developing a hateful memes dataset consisting of around 10K synthetic memes labeled in hateful and not-hateful classes. Similarly, another large-scale dataset comprising 150K memes was introduced by \newcite{gomez2020exploring} for hate speech detection. In another work, \newcite{suryawanshi2020multimodal} developed an offensive memes dataset comprising 743 memes collected during the event of the 2016 US presidential election. \newcite{pramanick2021detecting} built a dataset to detect harmful memes containing around 3.5K related to COVID-19. 

Some works have also been conducted concerning low-resource languages. \newcite{perifanos2021multimodal} developed a dataset of 4,004 memes for detecting hate speech in Greek. 
\newcite{kumari2023emoffmeme} developed an offensive memes dataset consisting of 7,417 Hindi memes. Recently, \newcite{das2023banglaabusememe} introduced a dataset comprising 4,043 samples for detecting the Bengali abusive memes. Two prior studies focused on hateful meme detection in the Bengali language. \newcite{karim2022multimodal} introduced a synthetic hate speech dataset comprising around 4,500 Bengali memes. Similarly, \newcite{hossain-etal-2022-mute} developed another Bengali hateful memes dataset having 4,158 memes labeled hateful or not-hateful. However, none of these datasets annotated the targets of hateful memes.

\noindent
\textbf{Multimodal hateful meme detection:} Over the years,  various approaches have been employed for detecting hate speech using multimodal learning. Earlier researchers used conventional fusion (i.e., early and late) \cite{suryawanshi2020multimodal,gomez2020exploring} produce producing multimodal representation. Later, some researchers employed bilinear pooling \citep{chandra2021subverting} while others developed transformer-based multimodal architectures such as MMBT \cite{kiela2019supervised}, ViLBERT \cite{lu2019vilbert}, and Visual-BERT \cite{li2019visualbert}. Besides, some works attempted to use disentangled learning \citep{lee2021disentangling}, incorporate additional caption \citep{zhou2021multimodal}, and add external knowledge \cite{pramanick-etal-2021-momenta-multimodal} to improve the hateful memes detection performance. Recently, \newcite{cao-etal-2022-prompting} applied prompting techniques for hateful meme detection in English.

\noindent
\textbf{Differences with existing studies:} Though many studies have been conducted on hateful meme detection, only a few studies have focused on Bengali.  We point out several drawbacks in the existing research on Bengali. Firstly, the existing hateful memes datasets are small. They framed the task as a binary (hateful or not-hateful) classification problem, overlooking the social entities (i.e., individuals, society) that a hateful meme can target. Our dataset provides both levels of annotation. The richness of our dataset contributes to a comprehensive understanding of the dynamics of Bengali memes. Only one particular work \cite{das2023banglaabusememe} studied the targets of abusive memes in Bengali. Their work is completely different from ours as we attempt to identify targeted social entities of hateful memes, which are more explicit than abuse. 
Secondly, the current research overlooked the memes containing cross-lingual captions, while many internet memes are written in code-mixed and code-switched manner. Lastly, none of the works provided any model that can be generalized across low-resource datasets.

\section{BHM: A New Benchmark Dataset}
As per our exploration, no dataset in Bengali currently focuses on capturing hateful memes that target specific entities. To fill this gap, we develop a novel multimodal hateful meme dataset. We followed the guidelines outlined by \newcite{hossain-etal-2022-mute} and \newcite{kiela2020hateful} to develop the dataset. This section will provide a detailed discussion of the dataset development steps, including the data collection and annotation process and relevant statistics.

\subsection{Data Collection and Sampling}
We collected memes from various online platforms, including Facebook, Instagram, Pinterest, and Blogs. We used keywords such as \textit{"Bengali Troll Memes", "Bengali Faltu Memes", "Bengali Celebrity Memes", "Bengali Memes", "Bengali Funny Memes", "Bengali Political Memes"}, etc to search for these memes. To avoid copyright infringement, we only collect the memes from publicly accessible pages and groups. We accumulated a total of 7,532 memes from March 2022 to April 2023. The distribution of data sources is presented in Appendix \ref{source}. We collected the memes with Bengali and code-mixed (Bengali + English) embedded texts. During the data collection, we filtered out memes if they (i) contained only visual or textual information, (ii) contained drawings or cartoons, and (iii) had uncleared contents either visually or textually. Appendix Figure \ref{filter} presents some filtered meme samples. We also removed duplicate memes. After filtering and deduplication, we discarded 299 memes and ended up with a curated dataset of 7,233 memes. Following these, we use an OCR library (PyTesseract\footnote{https://pypi.org/project/pytesseract/}) to extract captions from the memes. As the captions have code-mixed texts, we manually reviewed and corrected if there were any missing words or spelling errors in the extracted captions. Finally, the memes and their captions are passed for manual annotation.   

\subsection{Dataset Annotation}
We develop \textbf{BHM} focusing on two tasks: (i) detecting whether a meme is hateful or not and (ii) identifying the targeted entity of a hateful meme. We create guidelines defining the tasks to ensure the annotation quality and mitigate the bias. Appendix Figure \ref{data-sample} depicts a few annotated meme examples.

\subsubsection{Definition of Categories}
Following the definition of previous studies \citep{kiela2020hateful,hossain-etal-2022-mute}, we consider \textit{a meme as hateful if it explicitly intends to denigrate, vilify, harm, mock, abuse any entity based on their gender, race, ideology, belief, social, political, geographical and organizational status}. Moreover, we define four target categories of hateful memes that the annotators can adhere to during annotation. The four target categories are as follows\footnote{The definitions reference individuals and organizations are based in Bangladesh.}:

\begin{enumerate}
    \item \textbf{Targeted Individual (TI)}: The hate directed towards a specific person (male or female) based on his fame, gender, race, or status. The person might be an artist, an actor, or a well-known politician such as \textit{Sakib Khan, Mithila, Khaleda Jia, Sajeeb Wazed, Tamim Iqbal} etc.    
    
    \item \textbf{Targeted Organization (TO)}: When the hate propagates towards any particular organization which is a group of people having certain goals such as a business company (e.g., \textit{Grameenphone, Airtel}), government institution(e.g., \textit{ School \& Colleges}), political organization (e.g., \textit{BNP, Awami League}), etc.
    
    \item \textbf{Targeted Community (TC)}: Hate on any specific group of people who hold common beliefs or ideology towards any religion (e.g., \textit{who follow the ideology of Buddhism}), culture (e.g., \textit{who celebrates the Bengali Pohela Baisakh or valentines day}), person (e.g., \textit{followers of cricketer Tamim Iqbal}) or organization (e.g., \textit{followers of Bangladesh National Party}).
    
    \item \textbf{Targeted Society (TS)}: When a meme promotes hate towards a group of people based on their geographical areas such as mocking entire \textit{Indian people} or \textit{British people} it becomes hateful to an entire society. 
\end{enumerate}

\subsubsection{Annotation Process}
To carry out the annotation process, we employed six annotators: four were undergraduate students, and two were graduate students, falling within the age range of 23 to 27 years. The group comprised four male and two female annotators, each possessing prior research experience in the field of NLP. Furthermore, to resolve any disagreement among the annotators, we included an expert with 15 years of experience in NLP. We divided the annotators into three groups of two people, each annotating different subsets of memes. Initially, we trained our annotators with the definition of hateful memes, their categories, and associated samples. Our primary goal was to ensure that the annotators comprehended the guidelines and identify hateful memes and their target entities. 

Two annotators independently annotated each meme, and the final label was determined through consensus between the annotators. On average, an annotator spent 3 minutes deciding the label of a meme. In disagreement, an expert provided the final decision by discussing the uncertainties. For a minimal number of memes (< 2\%), we observe that memes target multiple entities. Since this number is minimal for annotation simplicity, such samples were annotated with the dominant class label. During the final label assignment, we discarded 85 memes as the annotators, and the expert could not agree on assigning a label. Finally, we get BHM, a multimodal Bengali hateful memes dataset with their targeted entities containing 7,148 memes.    
\begin{table}[h!]
\begin{center}
\small
\begin{tabular}{llcc}
\hline
 &  \textbf{Label} & \textbf{$\kappa$-score}  & \textbf{Average}\\
\hline
\multirow{2}{*}{\begin{tabular}[c]{@{}c@{}}Task 1\end{tabular}}& Hate & 0.82 &\multirow{2}{*}{0.79} \\
& Not Hate & 0.76 & \\
\hline
\multirow{4}{*}{\begin{tabular}[c]{@{}c@{}}Task 2\end{tabular}}& Target Individuals & 0.68  & \multirow{4}{*}{0.63} \\
& Target Organizations & 0.66 &    \\
& Target Communities & 0.61 &      \\
& Target Society  & 0.57 &    \\
\hline 
 
\end{tabular}
\caption{Cohen's $\kappa$ agreement score during the annotation of each task: Task 1: hateful meme detection (2-class classification) and Task 2: target identification (4-class classification) of hateful memes.}
\label{kappa-score}
\end{center}
\end{table}

\noindent
\textbf{Inter-annotator Agreement: } We computed the inter-annotator agreement in terms of Cohen's Kappa ($\kappa$) score \cite{Cohen1960} to check the validity.  Table 1 shows the Kappa scores for each task category. We achieved a high agreement score of 0.79 for the hateful meme detection task. However, for target identification, we attained a moderate score of 0.63. These agreement scores indicate that annotators struggled distinguishing the targeted entities within hateful memes.

\begin{figure*}[!b]
    \centering
    \includegraphics[width =0.7\textwidth]{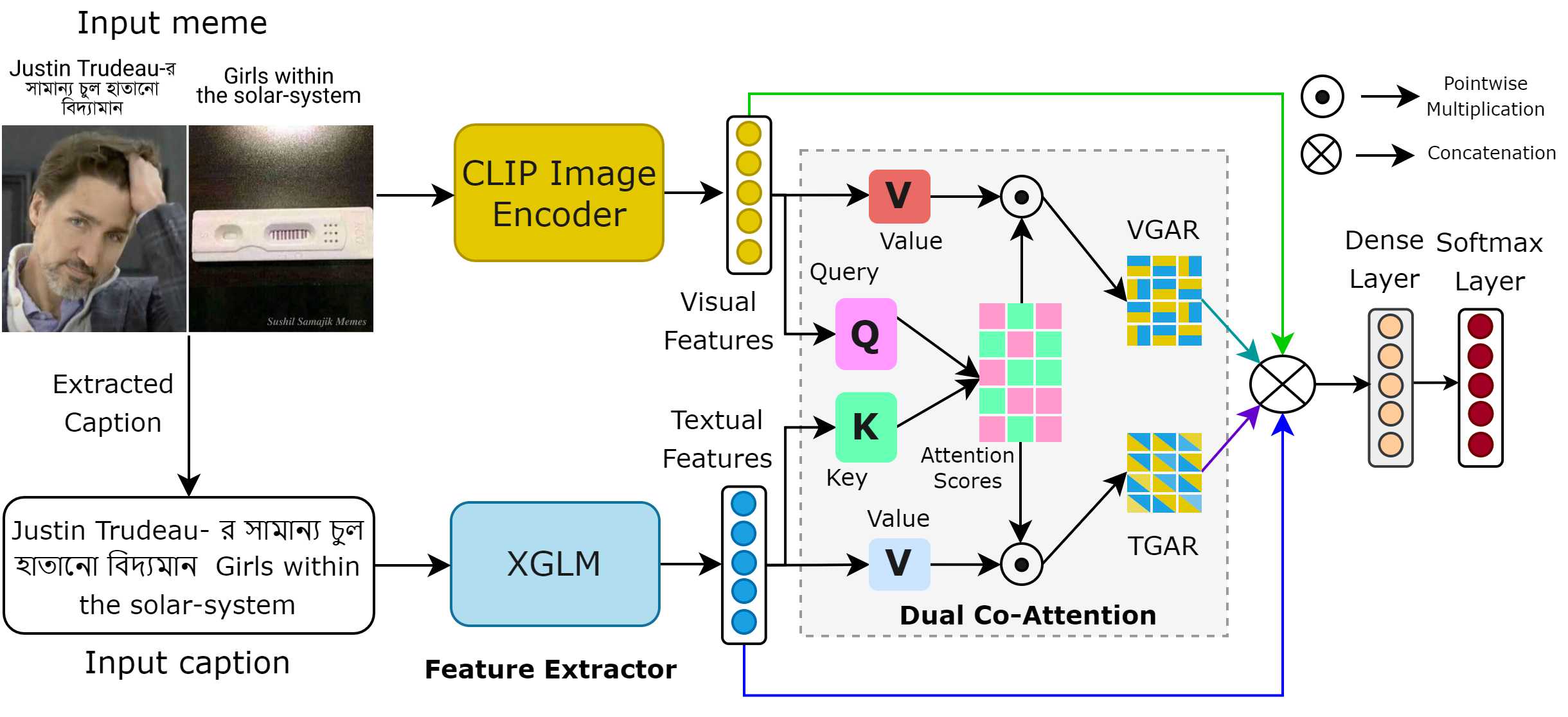}
    \caption{A simplified view of our proposed Dual Co-Attention Framework ({\fontfamily{qcr}\selectfont DORA}). The upper block represents the visual feature extractor, and the lower block is the textual feature extractor. The Dual Co-Attention block takes encoded visual and textual representation and generates two attentive vectors: VGAR (Vision-guided attentive Representation) and TGAR (Text-guided Attentive Representation). Finally, our method generates a richer multimodal representation by concatenating the attentive vectors with the individual modality-specific features.}
    \label{method}
\end{figure*}

\subsection{Dataset Statistics}
We divided the dataset into training (80\%), validation (10\%), and test (10\%) sets for model training and evaluation. Table \ref{split} shows the data distribution across different categories within each split. Task 1 exhibits a slight imbalance, while task 2 presents a significant imbalance, with most data falling under the `TI' category. The distribution highlights that it will be challenging to accurately identify the targeted entities in hateful memes due to the limited number of samples in the `TO', `TC', and `TS' categories. We analyze the training set memes to acquire more insights into data characteristics. The analysis is presented in Appendix \ref{data-stat}.

\begin{table}[h!]
\begin{center}

\small
\begin{tabular}{l|lccccc}
\hline
 &\textbf{Class} & \textbf{Train} & \textbf{Valid} & \textbf{Test} & \textbf{Total}\\
\hline
\multirow{2}{*}{Task 1}  & HT & 2117  & 241 & 266 & 2624 \\
& NHT & 3641  & 399 & 445 & 4485 \\
\hline
\multirow{4}{*}{Task 2}  & TI & 1623 & 192 & 193 & 2008 \\
&TO & 160 & 17 & 27 & 204  \\
&TC & 249 & 24 & 37 & 310     \\
&TS  & 85 & 8 & 9 & 102    \\
\hline 
\end{tabular}
\caption{Number of memes in train, test, and validation set for each category.}
\label{split}
\end{center}
\end{table}

\section{Methodology}
This section describes the proposed multimodal framework for detecting hateful memes and their targeted entities. 
Figure \ref{method} shows the overall architecture of the proposed system.       

\subsection{Feature Extractor}
To encode the visual information of the memes, we leverage the image encoder component of the CLIP (Contrastive Language Image Pretraining) \cite{radford2021learning}, a prominent visio-linguistic model. This image encoder incorporates a vision transformer \cite{dosovitskiy2020image} as its backbone. 
Meanwhile, we leverage the XGLM \cite{lin2022few}, a multilingual generative language model, to encode meme captions. XGLM has effectively learned from diverse languages within context without parameter updates. 
Given that our dataset comprises code-mixed captions (Bangla + English), we posited that XGLM could offer a better contextualized representation of these code-mixed captions. We fine-tuned both the image and text encoders, aiming to extract encoded representations. These representations were subsequently fed into a dual co-attention module to produce a multimodal representation.

\subsection{Dual Co-Attention}
We fed the encoded visual and textual features into a dual co-attention block to generate an effective multimodal representation. Specifically, we generate two attentive multimodal feature representations using the Multi-head Self Attention (MSA) mechanism. The MSA takes three matrices: Query (Q), Key (K), and Value (V) as input. In standard NLP applications, all the matrices come from word representations. In contrast, motivated by \newcite{lu2019vilbert}, we modified the MSA block where queries come from one modality and keys and values from another. This modification will generate an attention-pooled representation for one modality conditioned on another. 

Specifically, in our case, the Q will be generated from visual features, whereas the K will be from textual features. Afterward, to determine the similarity between the visual and textual features, we calculated the attention values by performing a dot product between Q and K. We then generated two different Value matrices, one coming from visual features and another from textual features. The goal was to generate two attentive representations where one modality guided another.

To do this, we first weighed the visual features by performing a point-wise multiplication with attention scores and named it vision-guided attentive representation (VGAR). 
After observing the visual information, we also generate another attentive representation by weighing the textual features. We called this text-guided attentive representation (TGAR). These two attentive representations (VGAR and TGAR) now contain significant cross-modal features. This cross-modal representation is further concatenated with the individual modality features (obtained from CLIP and XGLM). This process will boost the gradient flow and help the model learn from the individual and their cross-modal features. Finally, the combined multimodal representation is passed to the dense layer, followed by a softmax operation to predict the meme's category.                   
\section{Experiments}
This section discusses the baselines and their performance comparison with the proposed method ({\fontfamily{qcr}\selectfont DORA}) and its variants. 
We developed several state-of-the-art computational models, including only visual, textual, and multimodal models pre-trained on both modalities. We used weighted F1 scores as the primary evaluation metrics. Other metrics, such as precision and recall, are also reported for the comparison. Appendix \ref{es} provides the details of the experimental settings.  

\subsection{Baselines}
We implemented several baseline models that were proven superior in similar multimodal hateful meme detection studies \cite{pramanick-etal-2021-momenta-multimodal, hossain-etal-2022-mute, sharma2023characterizing}.


\subsubsection{Unimodal Models}
For the visual-only models, we employed three well-known architectures: \textbf{Vision Transformer (ViT)} \cite{dosovitskiy2020image}, \textbf{Swin Transformer (SWT)} \cite{liu2021swin}, and \textbf{ConvNeXT} \cite{liu2022convnet}. Three pre-trained textual-only transformer models, namely \textbf{Bangla-BERT} \cite{Sagor_2020}, \textbf{multilingual BERT} \cite{devlin-etal-2019-bert}, and \textbf{XLMR} \cite{conneau-etal-2020-unsupervised} are used. We fine-tuned the unimodal models on the developed dataset. 

\subsubsection{Multimodal Models}
\begin{itemize}

    \item \textbf{MMBT}: Multimodal BiTransformer (MMBT) \cite{kiela2019supervised} uses a transformer \cite{vaswani2017attention} architecture for fusing the visual and textual information.
   
    \item \textbf{CLIP}: It is a multimodal model trained on noisy image-text pair using contrastive learning \cite{chen2020simple} approach. CLIP has been widely used for several multimodal classification tasks \cite{pramanick2021momenta,kumar2022hate}. We extract the visual and textual embedding representations by fine-tuning the CLIP on the developed dataset. Afterward, we combined both representations and trained them on top of a softmax layer. 
    
    \item \textbf{ALBEF}: ALBEF (Align Before Fuse) \cite{li2021align} is another state-of-the-art multimodal model that uses momentum distillation and contrastive learning method for the pre-training on noisy image-text data.
\end{itemize}

\begin{table*}[t!]
\centering

\scriptsize
\begin{tabular}{ll|ccc|ccc}

\hline
\textbf{Approach}&\textbf{Models} & \multicolumn{3}{c}{\textbf{Hateful Meme Detection (Task 1)}}& \multicolumn{3}{c}{\textbf{Target Identification (Task 2)}}\\
\hline
&&\textbf{P}&\textbf{R}&\textbf{F1}& \textbf{P}& \textbf{R}&\textbf{F1}\\
\hline                   
\multirow{3}{*}{\textbf{Visual Only}}
 & ViT & 0.677 & 0.682 & 0.645  & 0.704 & 0.631 & 0.645 \\
 & SWT &  0.669 & 0.680 & 0.660 & 0.682 & 0.620 & 0.645  \\
 & ConvNeXT &  \underline{0.692} & \underline{0.699} & 0.665 & 0.604 & 0.650 & 0.622 \\
 \midrule  
\multirow{3}{*}{\textbf{Text Only}} 
 & Bangla BERT & 0.644 & 0.658 & 0.644 & 0.634 & 0.597 &	0.612 \\
 & mBERT & 0.628 & 0.648 & 0.610 & \underline{0.706} & 0.616 & 0.652 \\
 & XLMR & 0.640 & 0.655 & 0.638  & 0.555 & 0.672 & 0.601  \\

\midrule          

\multirow{3}{*}{\textbf{Multimodal}}
  
& MMBT & 0.629 & 0.646 & 0.587 & 0.704 & 0.657 & 0.662  \\ 
& CLIP & 0.596 & 0.607 & 0.600  & 0.550 & 0.714 & 0.617  \\ 
& ALBEF & 0.671 & 0.682 & \underline{0.670} & 0.649 & \underline{0.740} & \underline{0.668}  \\ 
\midrule
\multirow{7}{*}{\begin{tabular}[c]{@{}c@{}}\textbf{Proposed System}\\ \textbf{and Variants}\end{tabular}}
& {\fontfamily{qcr}\selectfont DORA} w/o VF & 0.692 & 0.697 & 0.693  & 0.592 & 0.736 & 0.644  \\  
& {\fontfamily{qcr}\selectfont DORA} w/o TF & 0.694 & 0.696 & 0.694  & 0.647 & 0.751 & 0.664 \\
& {\fontfamily{qcr}\selectfont DORA} w/o VF+TF & 0.694 & 0.689 & 0.691 & 0.536 & 0.725 & 0.616 \\
& {\fontfamily{qcr}\selectfont DORA} w/o VGAR & 0.688 & 0.696 & 0.675  & 0.639 & 0.740 &0.679  \\
& {\fontfamily{qcr}\selectfont DORA} w/o TGAR & 0.672 & 0.682 & 0.654  & 0.659 & 0.729 & 0.677  \\
& {\fontfamily{qcr}\selectfont DORA} w/o VGAR + TGAR & 0.693 & 0.696 & 0.665  & 0.662 & 0.744 & 0.686  \\
& {\fontfamily{qcr}\selectfont DORA} & \textbf{0.718} & \textbf{0.718} & \textbf{0.718} & \textbf{0.706} & \textbf{0.759} & \textbf{0.720} \\
\hline
\multicolumn{2}{c|}{$\Delta_{{\fontfamily{qcr}\selectfont DORA}-baseline\_model}$}  & \textcolor{blue}{2.6} & \textcolor{blue}{1.9} & \textcolor{blue}{4.8} & \textcolor{blue}{0} & \textcolor{blue}{1.9} & \textcolor{blue}{5.2}\\
\hline
\end{tabular}
\caption{\label{result:1} Performance comparison of visual only, textual only, and multimodal models on the test set. P, R, and F1 denote precision, recall, and weighted F1-score, respectively. The VF, TF, VGAR, and TGAR denote the visual features, textual features, vision-guided attentive representation, and text-guided attentive representation. The best-performing score in each column is highlighted in \textbf{bold}, and the second-best score is \underline{underlined}. The last row shows the performance improvement of the proposed system ({\fontfamily{qcr}\selectfont DORA}) over the best baseline score. 
}
\end{table*}

\begin{table}[t!]
\centering

\scriptsize
\begin{tabular}{ll|ccccc}

\hline
\textbf{Model}&\textbf{Categories} & \textbf{P}&\textbf{R}&\textbf{F1}& \textbf{Ma.F1}& \textbf{W.F1}\\
\hline                   
\multirow{4}{*}{\textbf{ALBEF}}
 & TI & 0.77 &  0.98 & 0.86  & \multirow{4}{*}{0.32}&  \multirow{4}{*}{0.67}\\
 & TC &  0.11  &  0.03  & 0.04 &  &  \\
 & TO &  0.78  & 0.26 & 0.39 &  &  \\
 & TS &  0.00 & 0.00 & 0.00 &  &  \\
 \midrule  
\multirow{4}{*}{\textbf{{\fontfamily{qcr}\selectfont DORA}}}
 & TI & 0.82  &  0.94 & 0.87  & \multirow{4}{*}{\textbf{0.45}}&  \multirow{4}{*}{\textbf{0.72}}\\
 & TC &  0.29 & 0.11  & 0.16 &  &  \\
 & TO &  0.55 & 0.59 & 0.57 &  &  \\
 & TS &  0.50 & 0.11 & 0.18 &  &  \\
\hline
\end{tabular}
\caption{\label{result:2} Class-wise performance comparison of the best model with the {\fontfamily{qcr}\selectfont DORA}. Here, Ma.F1 and W.F1 indicate macro and weighted F1-score respectively.
}
\end{table}

\subsection{Results}
Table \ref{result:1} shows the performance of all the models for hateful meme detection and its target identification.

\paragraph{Hateful Meme Detection:} Among the visual approach, the ConVNeXT model obtained the highest score (F1: 0.665). In the case of the textual-only models, Bangla BERT outperformed others (mBERT, XLMR) with an F1 score of 0.644. Notably, this score falls approximately 2.1\% short of the best visual model performance, indicating that visual information is more distinguishable in identifying hateful memes. In contrast, ALBEF surpassed both unimodal counterparts in the state-of-the-art multimodal model, attaining the highest F1 score of 0.670. MMBT and CLIP failed to deliver satisfactory results.
However, despite ALBEF's notable performance, our proposed method ({\fontfamily{qcr}\selectfont DORA}) outperforms the best model with an absolute improvement of 4.8\% in F1 score.

\paragraph{Target Identification:} 
Table \ref{result:1} illustrates that task 2's textual model (mBERT) achieved the best F1 score of 0.652 among unimodal models. Nonetheless, consistent with our earlier findings, the joint evaluation of multimodal information led to significant enhancements in target identification. In the case of the multimodal models, the ALBEF surpassed the unimodal counterparts, achieving the highest score of 0.668. However, {\fontfamily{qcr}\selectfont DORA} outperforms the best model by 5.2\% in terms of F1 score. 

To further illustrate the superiority of the {\fontfamily{qcr}\selectfont DORA}, we compared its class-wise performance with the best baseline model (ALBEF) presented in Table \ref{result:2}. ALBEF performs poorly across all the target categories, especially in the `TC' (0.04) and `TS' (0.0) classes. In contrast, {\fontfamily{qcr}\selectfont DORA} demonstrated significant improvement in F1 score across all the classes. Overall, {\fontfamily{qcr}\selectfont DORA} yielded an impressive 13\% improvement in the macro average F1 score, elevating it from 0.32 to 0.45. This improvement signifies that {\fontfamily{qcr}\selectfont DORA} maintains a good balance in the performance of all the evaluation measures (precision, recall, and F1 score).      
\begin{figure*}[t!]
  \centering
  \subfigure[\textbf{Visual:} TO (\xmark) \newline \textbf{Textual:} TS (\xmark)  \newline \textbf{{\fontfamily{qcr}\selectfont DORA}:} TI (\cmark)]{\includegraphics[scale=0.4,width =3.5cm,height = 2.5cm]{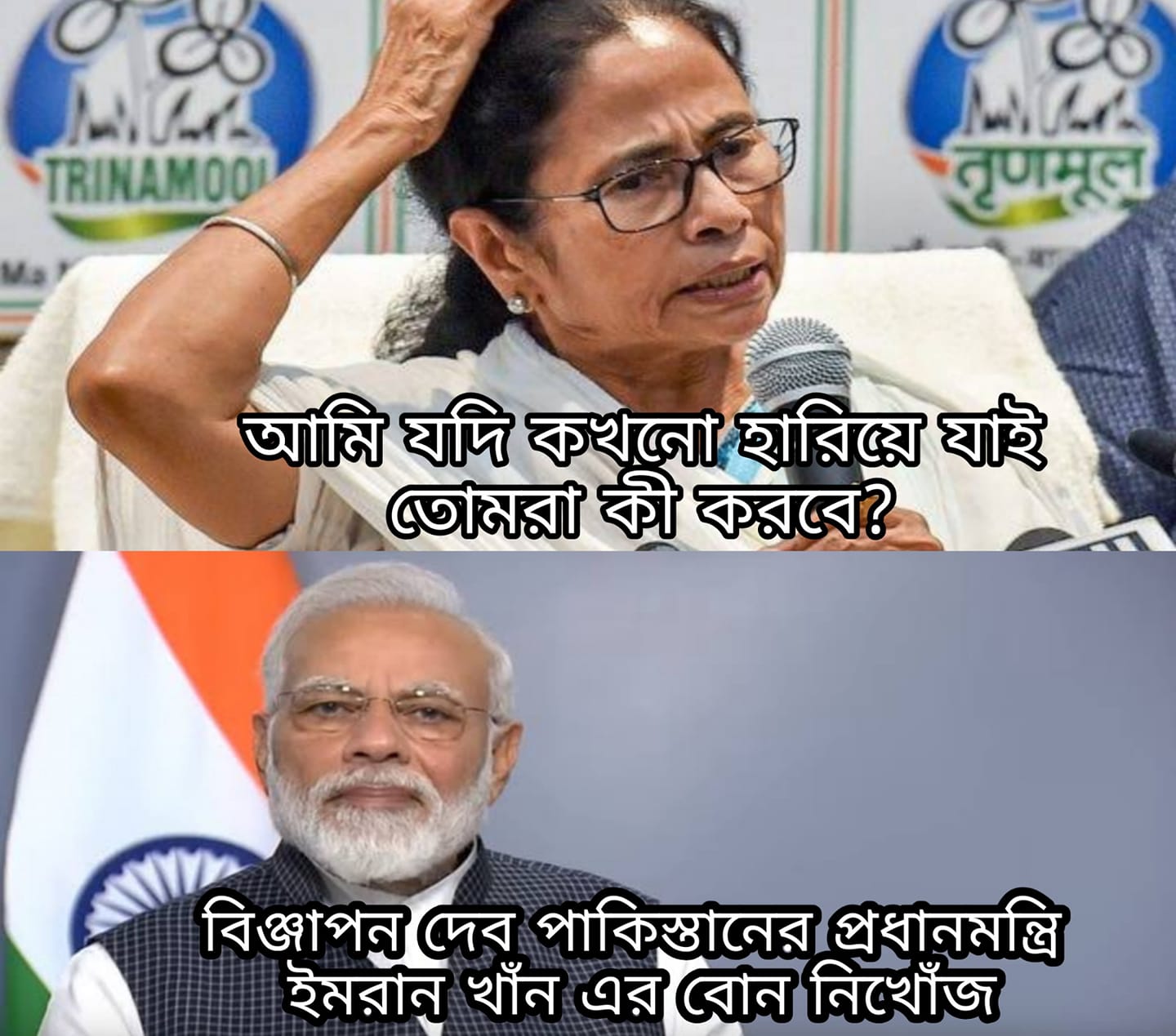}}\quad
  \subfigure[\textbf{Visual:} TI (\xmark) \newline \textbf{Textual:} TC (\xmark)  \newline \textbf{{\fontfamily{qcr}\selectfont DORA}:} TO (\cmark)]{\includegraphics[scale=0.4,width =3.5cm,height = 2.5cm]{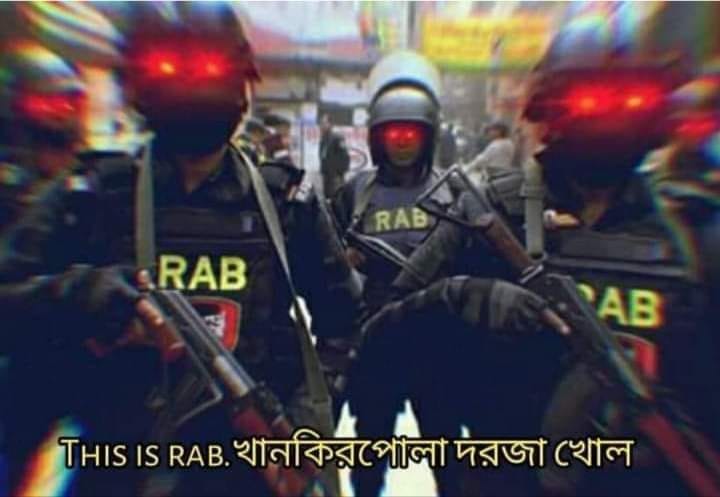}}\quad
  \subfigure[ \textbf{Actual:} TS  \newline  \textbf{Predicted:} \textcolor{red}{TI}]{\includegraphics[scale=0.4,width =3.5cm,height = 2.5cm]{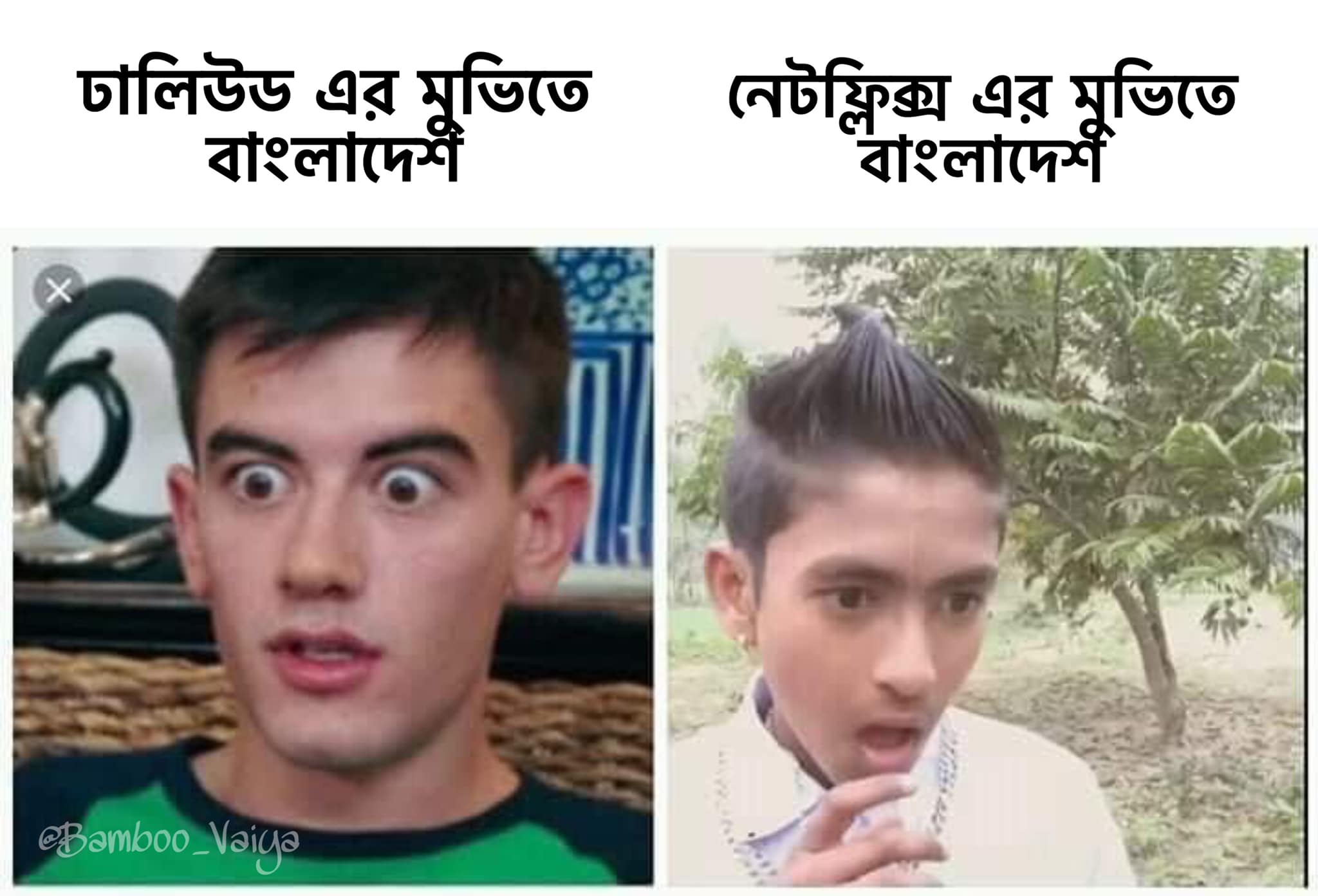}}

 \caption{Example (a) and (b) shows the memes where {\fontfamily{qcr}\selectfont DORA} yields better predictions, and example (c) illustrates a wrongly classified sample. The symbol (\cmark) and (\xmark) indicates the correct and incorrect prediction, respectively.} 
\label{qa}
\end{figure*}

\subsection{Ablation Study}
We perform an ablation study to analyze the contribution of each component (visual features (VF), textual features (TF), vision-guided attentive representation (VGAR), and text-guided attentive representation (TGAR)) of {\fontfamily{qcr}\selectfont DORA}. The last seven rows of Table \ref{result:1} show the ablation outcome.

For Task 1, it is noteworthy that even in the absence of VF and TF, the model's performance remains superior (0.691-0.694) compared to the best baseline model, ALBEF (0.670). However, a substantial drop in the F1 score (ranging between 0.654 to 0.675) occurs when the attentive representations are removed, underscoring the significant impact of the dual-co attention mechanism in our proposed approach. Conversely, in the case of target identification, the model exhibits diminished performance when the VF and TF are excluded. Interestingly, removing VGAR and TGAR shows less effect on the performance as the F1 score rises to its highest at 0.686. This implies that, for target identification, VF and TF bear greater significance than attentive representations. However, integrating all components in {\fontfamily{qcr}\selectfont DORA} results in a notable overall performance boost for both tasks.

\subsection{Transferability and Generalizability of {\fontfamily{qcr}\selectfont DORA}}

Table \ref{result:3} shows the transferability and generalizability of the {\fontfamily{qcr}\selectfont DORA} on two datasets (\textit{MUTE} \cite{hossain-etal-2022-mute} and \textit{EmoffMeme} \cite{kumari2023emoffmeme}) of different languages (i.e., Bengali and Hindi). Both \textit{MUTE} and \textit{EmoffMeme} have code-mixed captions in the memes, similar to the BHM dataset. To compare the performance in this experiment, we consider the best baseline model ( ALBEF). Results exhibit that when training and testing are done on the same dataset, {\fontfamily{qcr}\selectfont DORA} exceeds ALBEF by $\approx$ 4-5\% in terms of F1 score across all the datasets. This outcome illustrates that {\fontfamily{qcr}\selectfont DORA} can also generalize well across languages. Similarly, when trained on one dataset and tested on a different one, {\fontfamily{qcr}\selectfont DORA} yields a better score than ALBEF.

Interestingly, the model trained in the Hindi dataset (\textit{EmoffMeme}) exhibits poor performance when tested on Bengali datasets. Conversely, the model trained on the Bengali datasets and tested on the Hindi dataset exhibits suboptimal performance (0.64-0.65). This outcome emphasizes the need for a more sophisticated method that can be transferable across various languages. Overall, it can be stated that {\fontfamily{qcr}\selectfont DORA} is generalizable and also transferable across the datasets of the same languages. 

\begin{table}[t!]
\centering

\scriptsize
\begin{tabular}{ll|c|c|c}

\hline
& & \textbf{BHM} & \textbf{MUTE} & \textbf{EmoffMeme} \\
\hline
&&\textbf{F1}&\textbf{F1}&\textbf{F1}\\
\midrule                   
\multirow{2}{*}{\textbf{BHM (BEN)}}
 & ALBEF & 0.670 & 0.683 &0.583 \\
 & {\fontfamily{qcr}\selectfont DORA} &   \textbf{0.718} & \textcolor{blue}{0.744} & \textcolor{blue}{0.655} \\
 \midrule  
 \multirow{2}{*}{\textbf{MUTE (BEN)}}
 & ALBEF & 0.673 & 0.724 & 0.627 \\
 & {\fontfamily{qcr}\selectfont DORA} &  \textcolor{blue}{0.701} & \textbf{0.762} & \textcolor{blue}{0.647} \\
 \midrule 
 \multirow{2}{*}{\textbf{EmoffMeme (HIN)}}
 & ALBEF & 0.513 & 0.499 & 0.785 \\
 & {\fontfamily{qcr}\selectfont DORA} &  \textcolor{blue}{0.529} & \textcolor{blue}{0.493} & \textbf{0.824} \\
 \midrule 

\end{tabular}
\caption{\label{result:3} Transferability of best multimodal baseline and  ({\fontfamily{qcr}\selectfont DORA}) on two additional benchmark datasets namely \textit{MUTE} and \textit{EmoffMeme}. Here, BEN and HIN indicate the Bengali and Hindi languages, respectively. The models are trained on the dataset specified in the rows and tested on the dataset specified in the columns.  All the reported scores are weighted F1. The best transferable results are indicated in blue, and the scores in bold denote the best performance when models are trained and tested on the same dataset.    
}
\end{table}

\subsection{Error Analysis}
The results from table \ref{result:1} demonstrated that our proposed method {\fontfamily{qcr}\selectfont DORA} is superior in identifying the hateful memes and their targets compared to the unimodal counterparts. To gain insights into the model's mistakes, we conduct a qualitative error analysis by examining some correctly and incorrectly classified samples, as illustrated in Figure \ref{qa}. For better demonstration, we compare {\fontfamily{qcr}\selectfont DORA}'s prediction with the best visual (ViT) and textual model (mBERT) predictions specifically for task 2. 

In figure \ref{qa}(a), the visual model incorrectly identified the meme as belonging to the `Targeted Organization (TO)' class, likely due to the appearance of some famous political person faces. Simultaneously, the presence of a country name may lead the textual model to consider the meme as from the `Targeted Society (TS)' class. However, when both information is attended our proposed method correctly identified the meme as from the `Targeted Individual (TI)' hate category. Similarly, in figure \ref{qa} (b), the visual model labeled the meme as the `TI' category, and the textual model identified it as the `(TC)' class. The presence of multiple persons in visual information and slang words in textual information might have contributed to the misclassification. However, when visual and textual cues were jointly interpreted, the proposed method {\fontfamily{qcr}\selectfont DORA} correctly predicted the meme as `TO'. Nevertheless, there were instances where {\fontfamily{qcr}\selectfont DORA} failed to provide the correct outcome. For instance, in figure \ref{qa} (c), the meme belongs to the `TS' class and is misclassified by {\fontfamily{qcr}\selectfont DORA} as `TI'. This misjudgment may be attributed to inconsistent visual features, specifically the presence of two boys' faces, which could misleadingly suggest classification as `TI'. Incorporating world-level knowledge can help mitigate such model mistakes, which would be a promising avenue for future exploration.

\section{Conclusion}
This paper introduced a new large-scale multimodal dataset of 7,148 memes for detecting Bengali hateful memes and their targeted social entities. We also proposed {\fontfamily{qcr}\selectfont DORA}, a multimodal deep neural network for solving the tasks. Experiments on our dataset demonstrate the efficacy of {\fontfamily{qcr}\selectfont DORA}, which outperformed nine state-of-art baselines for two tasks. We further demonstrated the generalizability and transferability of {\fontfamily{qcr}\selectfont DORA} across other datasets of different languages. We plan to extend the dataset for more domains and languages in the future.

\section*{Limitations}
Though the proposed method ({\fontfamily{qcr}\selectfont DORA}) shows superior performance, there still exist some limitations of our work. First, we did not consider the background contexts, such as visual entities (i.e., detected objects) and textual entities (i.e., person name, organization name), as external knowledge to the model, which could improve the overall performance. Second, it is likely that in some cases, the {\fontfamily{qcr}\selectfont DORA} may focus on less significant parts of the content while attending to the information. If the dataset contains misleading captions or irrelevant textual information, the attention mechanism might align with those parts of the image that are visually unrelated, producing misleading representations. Incorporating adversarial training could be an interesting future direction to mitigate the generation of such biased multimodal representations. Third, we observed that our method {\fontfamily{qcr}\selectfont DORA} struggled with memes that convey hate implicitly. It appeared to have difficulty correctly interpreting cultural references and context-specific content, leading to additional incorrect predictions. We plan to address this aspect in the future. 


\section*{Ethical Considerations} 
\textbf{User Privacy:} All the memes in the dataset were collected and annotated in a manner consistent with the terms and conditions
of the respective data source. We do not collect or share any
personal information (e.g., age, location, gender identity)
that violates the user’s privacy.\\

\noindent
\textbf{Biases:} Any biases found in the dataset and model are unintentional. A diverse group of annotators labeled the data following a comprehensive annotation
guideline, and all annotations were reviewed to address any
potential annotation biases. We randomly collected data from various public social media pages and blogs to reduce data source biases. Moreover, we used neutral keywords (e.g., Bengali Memes, Bengali Mojar Memes, Funny Memes, Bengali Hashir Memes) not explicitly tied to specific hate themes to mitigate biases toward any specific person, community, or organization. Despite our best efforts, there may be inherent biases in the dataset, a common challenge in the dataset development process.\\

\noindent
\textbf{Intended Use:} We intend to make our dataset accessible to encourage further research on hateful memes. We believe this dataset will help in understanding and building models of low-resource, especially Asian languages.\\

\noindent
\textbf{Reproducibility:} We present the details of our experimental setting in Appendix A for the system's reproducibility. We will release the source code and the dataset at \url{https://github.com/eftekhar-hossain/Bengali-Hateful-Memes} upon accepting the paper.




\bibliography{custom}
\bibliographystyle{acl_natbib}

\appendix
\section*{Appendix}
\label{sec:appendix}
\counterwithin{figure}{section}
\counterwithin{table}{section}

\section{Experimental Settings}
\label{es}
For the experiment, we used the Google Colab platform. We downloaded the transformer models from the Huggingface\footnote{https://huggingface.co/} library and implemented it using the PyTorch Framework. The BNLP\footnote{https://github.com/sagorbrur/bnlp} and scikit-learn\footnote{https://scikit-learn.org/stable/} library has been used for the preprocessing and evaluation measures. We empirically selected the models' hyperparameter values by examining the validation set's performance. All the models are compiled using \textit{cross\_entropy} loss function. 

For optimizing the errors, in the case of the visual-only models, we used \textit{MADGRAD} \cite{defazio2022adaptivity} optimizer with a \textit{weight\_decay} of $0.01$. For task 1, we chose the learning\_rate $2e^{-5}$ while for task 2 it is settled to $7e^{-7}$. Conversely, for both tasks, among the textual-only models, mBERT and XLMR were trained using \textit{Adam} \cite{kingma2014adam} optimizer with learning rate $1e^{-5}$ while MADGRAD (learning\_rate = $2e^{-5}$ ) was utilized for Bangla-BERT model. Meanwhile, in the case of the multimodal models, ALBEF and CLIP were optimized using Adam (learning\_rate = $1e^{-5}$), and MMBT with MADGRAD (learning\_rate of $2e^{-4}$) optimizer. These settings of multimodal models were kept identical for both tasks.      

On the other hand, in the case of the proposed \textsc{DORA} and its variants, we use the two attention heads in the multi-head co-attention block. During training, the models were optimized using MADGRAD with a $2e^{-5}$ learning rate. We used the \textit{batch size} of 4 and trained the models for 20 \textit{epochs} with a learning rate scheduler. We examined the validation set performance to save the best model during training.

\section{Data Sources and Filtering}
\label{source}
\begin{figure}[!h]
    \centering
    \includegraphics[width=0.4\textwidth]{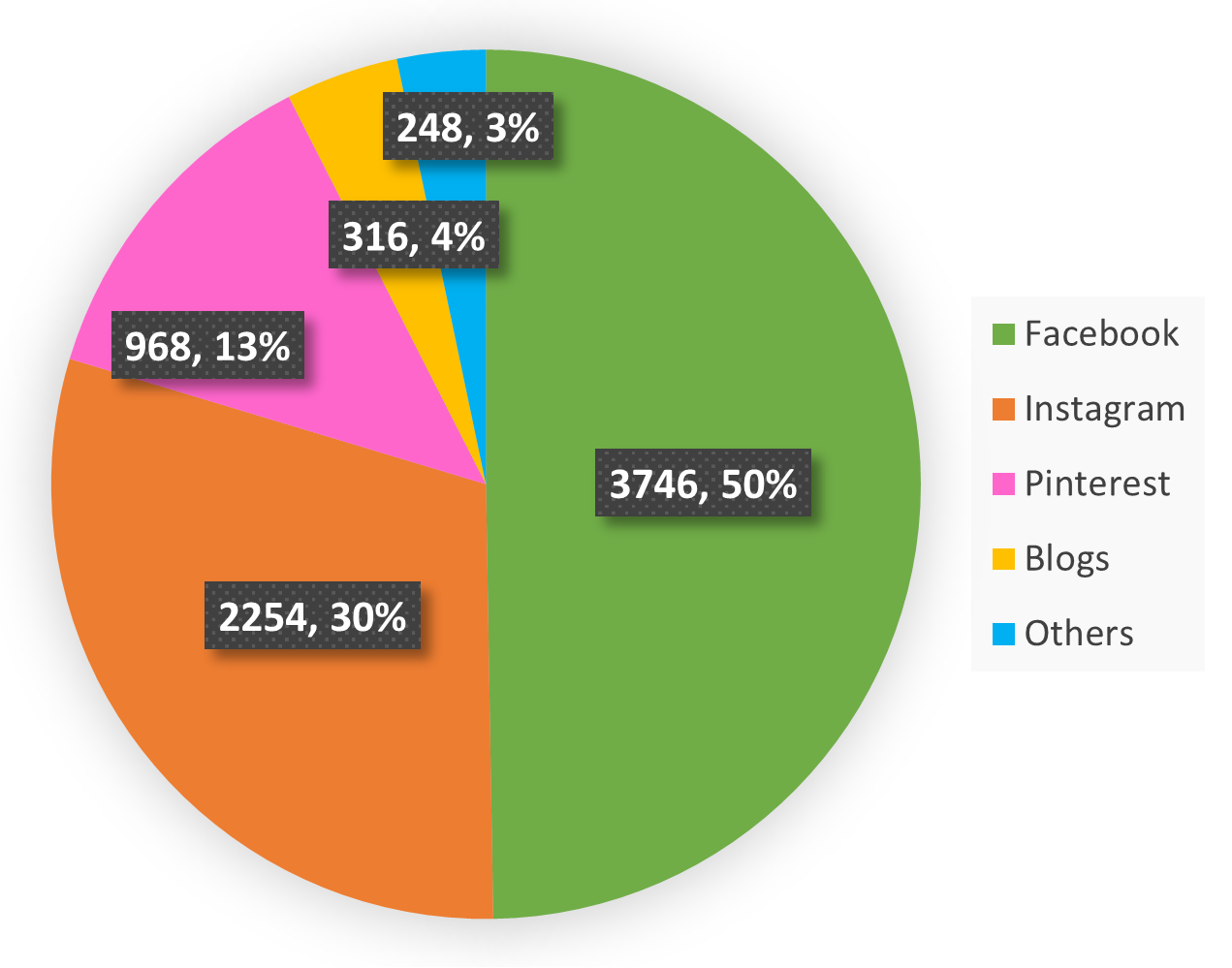}
    \caption{Distribution of data sources. Each cell represents the number and percentage of samples collected from the corresponding sources.}
    \label{fig:data_source}
\end{figure}

Figure \ref{fig:data_source} depicts the number of memes collected from each source. Most memes were collected from Facebook (50\%) and Instagram (30\%), while a few were accumulated from Pinterest, blogs, and other sources.
\begin{figure*}[]
  \centering
  \subfigure[Targets Individual]{\includegraphics[scale=0.45,width =3.5cm,height = 2.5cm]{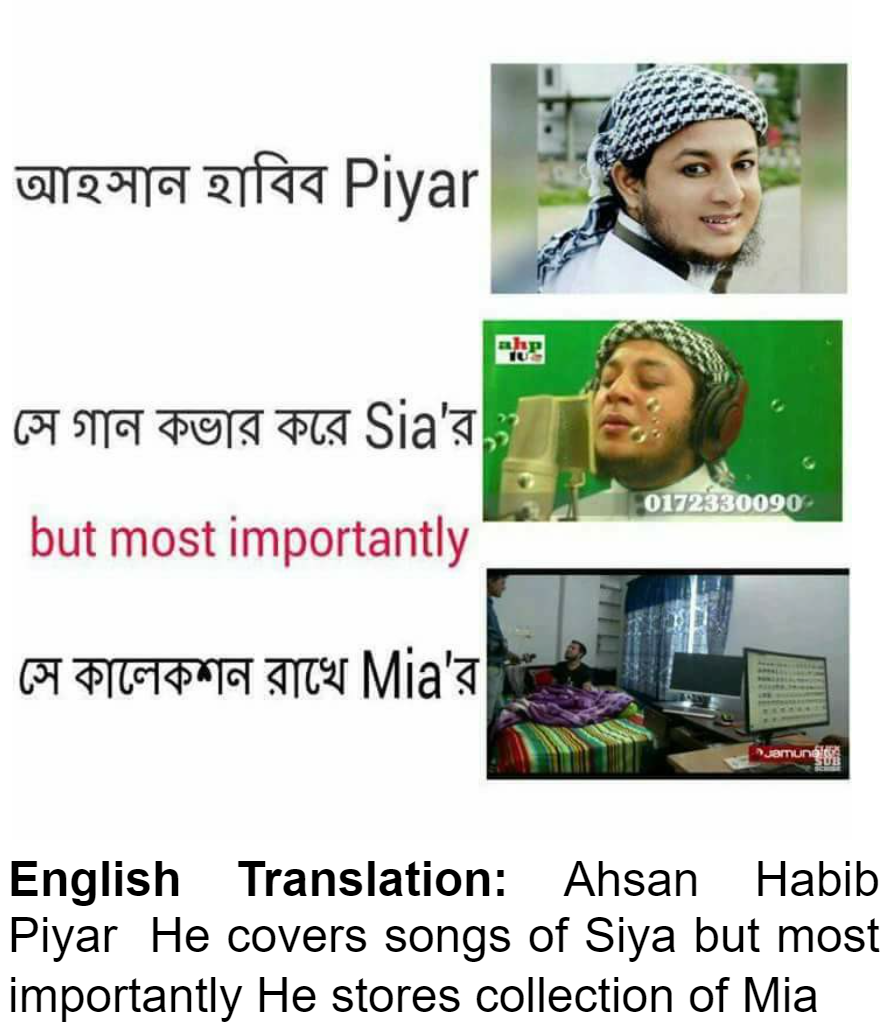}}\quad
  \subfigure[Targets Community]{\includegraphics[scale=0.45,width =3.5cm,height = 2.5cm]{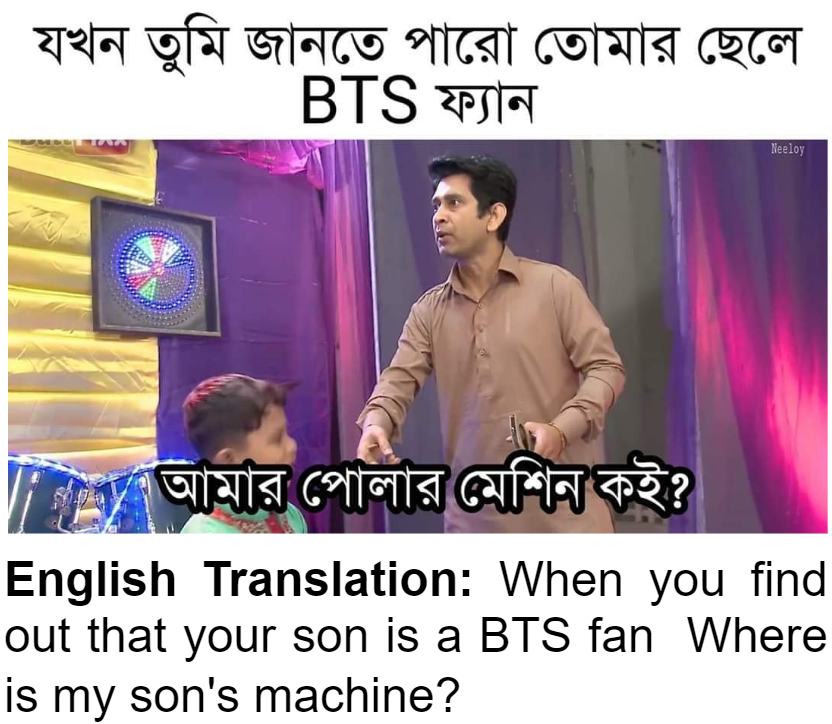}}\quad
  \subfigure[Targets Organization]{\includegraphics[scale=0.45,width =3.5cm,height = 2.5cm]{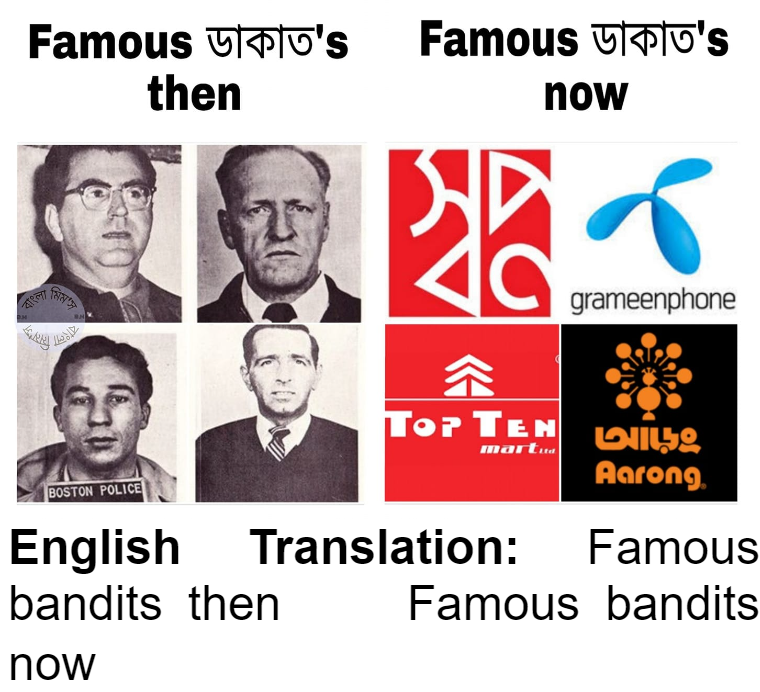}}
  \quad
  \subfigure[Targets Society]{\includegraphics[scale=0.45,width =3cm,height = 2.5cm]{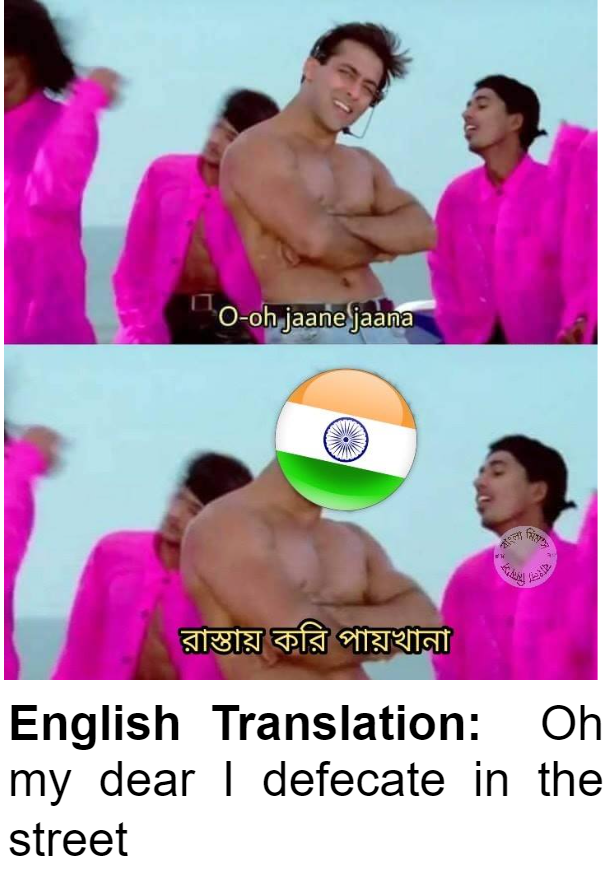}}

\caption{Few examples hateful memes targets from \textbf{BHM} dataset. The factors based on which the targets were decided (a) demean a person, (b) attack the sexual orientation of a community (BTS Fanbase), (c) state some organizations as Robbers, and (d) denigrate the people of a particular region.} 
 \label{data-sample}
\end{figure*}

\begin{figure}[h!]
  \centering
  \subfigure[]{\includegraphics[scale=0.4,width =3.5cm,height = 2.5cm]{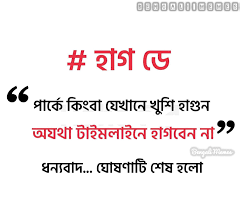}}\quad
  \subfigure[]{\includegraphics[scale=0.4,width =3.5cm,height = 2.5cm]{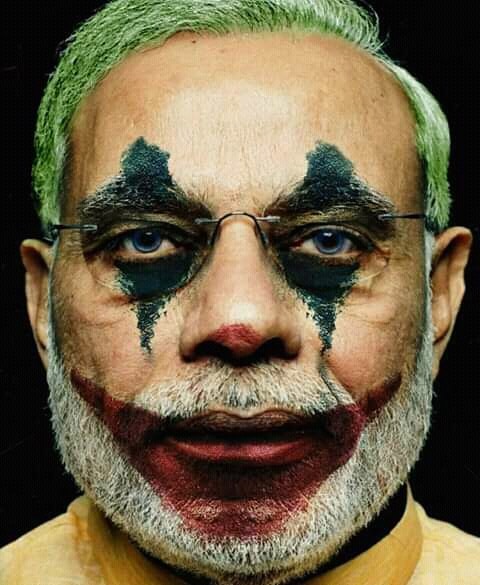}}\quad
  \subfigure[]{\includegraphics[scale=0.4,width =3.5cm,height = 2.5cm]{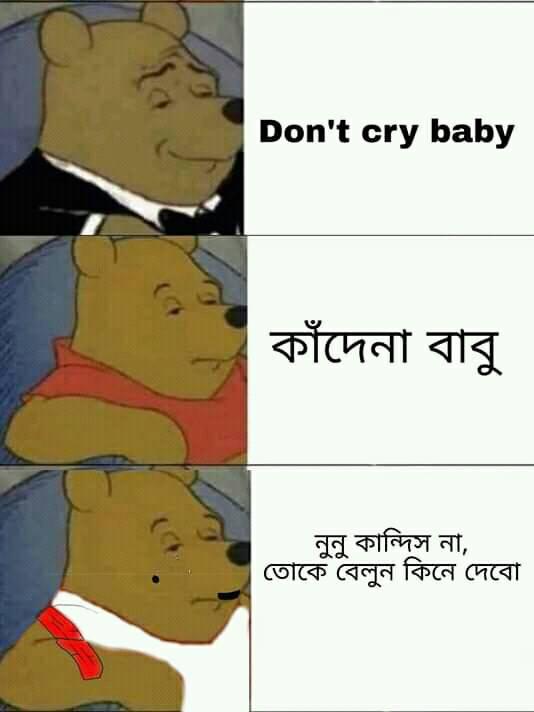}}
  \quad
  \subfigure[]{\includegraphics[scale=0.4,width =3.5cm,height = 2.5cm]{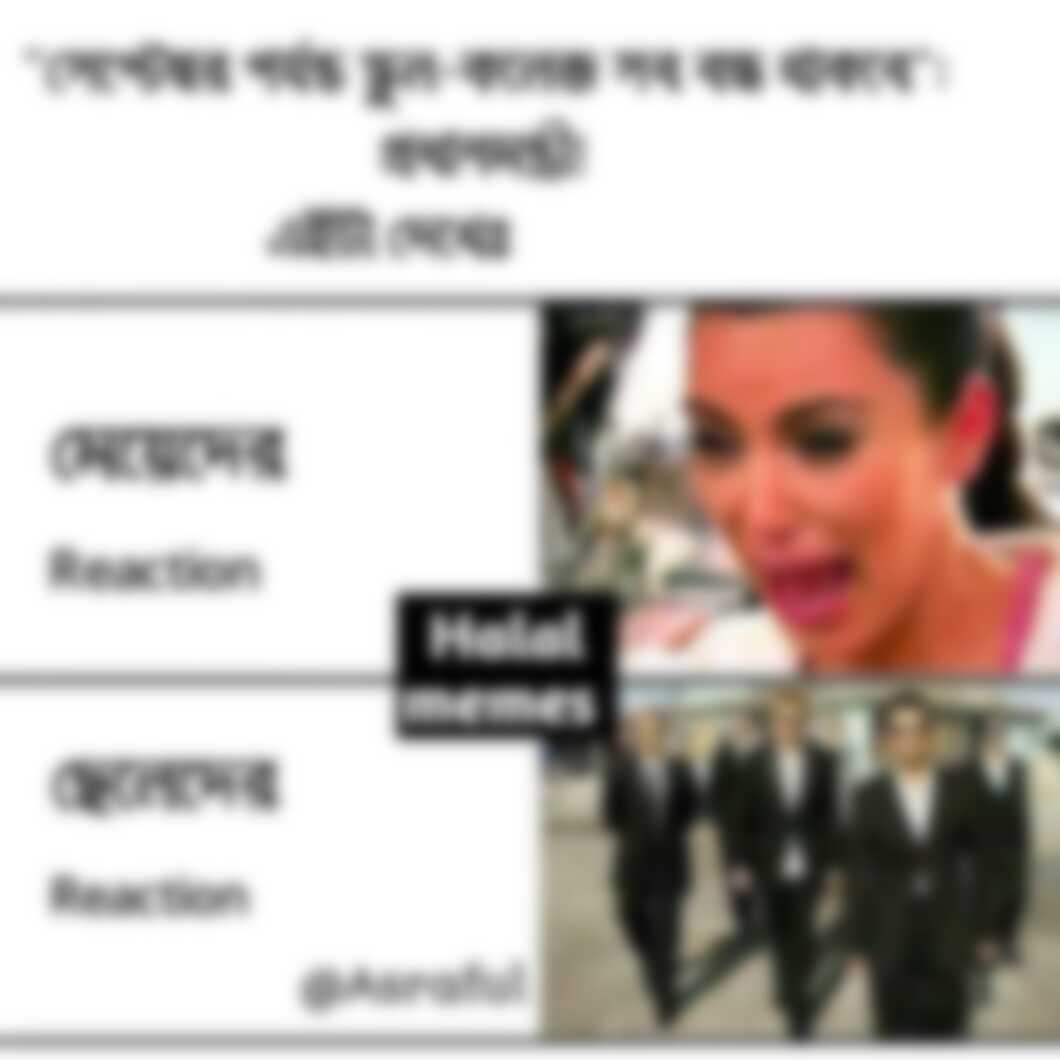}}

\caption{Example filtered memes during the data collection process.  The reason for the filtering is (a) contains only visual information (b) only textual information (c) contains cartoons (d) the contents are not cleared.} 
 \label{filter}
\end{figure}

\section{Additional Data Statistics}
\label{data-stat}
\begin{figure}[h!]
\centering
\includegraphics[width =\linewidth]{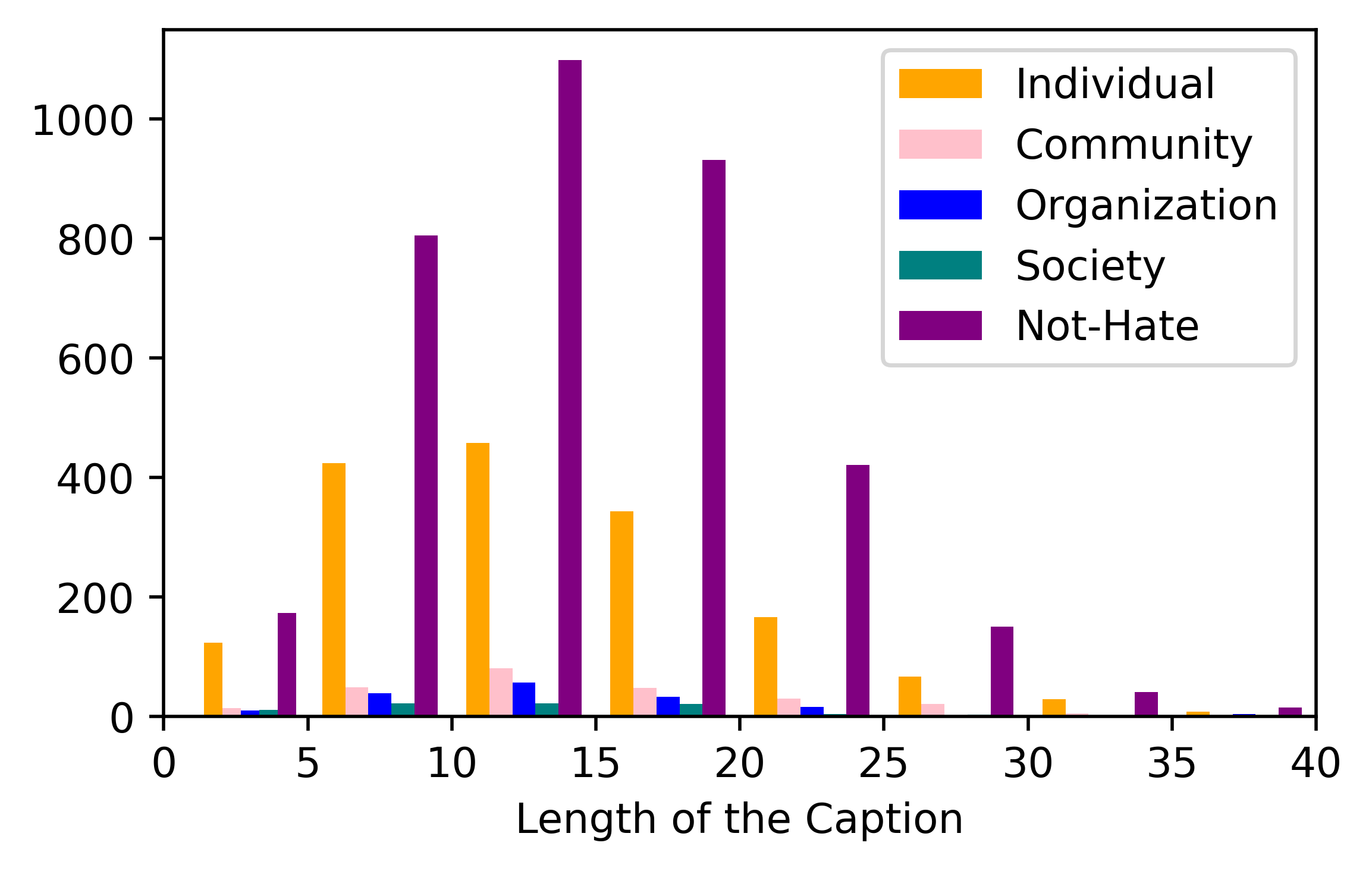}
\caption{Caption length distribution of the training set across different classes.} 
\label{hist}
\end{figure}

\paragraph{Text Analysis:} Table \ref{stats} presents lexical statistics for the training set meme captions. In Task 1, the NHT class exhibits the highest number of unique words(12,428) compared to the HT class (8,852). This discrepancy is unsurprising as the NHT class has the largest number of instances among all the classes. In the case of Task 2, the TI class got the highest count of unique words (7,150), while the TS class features the lowest (727). The average caption length remains consistent at 13 words for most classes, with exceptions in TC (14)  and TS (12). Figure \ref{hist} displays a histogram illustrating caption length across different classes. The distribution reveals that most captions fall within the 5 to 30-word range. Only the TI and NHT classes contain captions with lengths exceeding 30 words. We further analyzed the captions to quantify word overlap across different classes. Specifically, we computed the Jaccard similarity (JS) \cite{jain2017information} score between the top 400 most common words of different classes. The Jaccard score between each pair of classes is presented in Table \ref{jaccard}. We observed a substantial JS score of 0.51 between the HT and NHT classes, indicating a significant overlap in the words of these two classes. Regarding the target categories, the TI and TC pair exhibited the highest JS score (0.34), while the scores for the other categories remained below 0.20.

\begin{table}[h!]
\begin{center}
\small
\begin{tabular}{l|lcC{1.2cm}C{1.5cm}}
\hline
 &  \textbf{Class} & \textbf{\#Words}  & \textbf{\#Unique words} & \textbf{Avg. \#words/cap.}\\
\hline
\multirow{2}{*}{Task 1}& HT &  28477 & 8852 & 13.45  \\
& NHT & 50344 & 12428 & 13.82 \\
\hline
\multirow{4}{*}{Task 2} & TI & 21583 & 7150  & 13.29   \\
& TO & 2159 & 1362 & 13.49   \\
& TC & 3694 & 2017 & 14.83      \\
& TS  & 1041  & 727 & 12.24     \\
\hline 
 
\end{tabular}
\caption{Lexical analysis of captions in training set in terms of total words, total unique words, and average caption length across categories.}
\label{stats}
\end{center}
\end{table}

\begin{table}[h!]

\begin{center}
\small
\begin{tabular}{l|ccc}
\hline
   & \textbf{TO} & \textbf{TC} & \textbf{TS}  \\
\hline
TI   & 0.19  &  0.34 & 0.16  \\
TO   &  & 0.18 & 0.11   \\
TC   &  &      &  0.14    \\
\hline
& \textbf{NHT}& &   \\
\hline
HT  & 0.51 & &  \\

 \hline 
\end{tabular}
\caption{Jaccard similarity analysis among the top 500 common words across different class combinations.}
\label{jaccard}
\end{center}
\end{table}

\paragraph{Image Analysis: }  
The presence/absence of facial images is an important component in any meme. Therefore, we analyze the faces present in a meme. We employed the \textit{deepFace}\cite{serengil2021lightface}library to perform this analysis. It allowed us to determine whether a given meme contains any faces. If a face is detected, we also extract information regarding the associated gender and age. Our findings reveal that approximately 34.12\% of the memes contain no faces. Among this group, 33.28\% belong to the 'Hate' class. Of the total memes, 53.29\% feature male faces, with 35.3\% from the 'Hate' class and 64.64\% from the 'Not-Hate' class. Meanwhile, 12.53\% feature female faces, with 50.83\% within the 'Hate' class and 49.17\% within the 'Not-Hate' class. On average, the detected ages of males and females in the memes hover around 31 years. A key observation is the prevalence of male faces in the 'Hate' class, indicating that males are common targets of hateful content in the Bengali community.

\end{document}